\documentclass[a4paper, 10pt, conference]{ieeeconf}      
\usepackage{FG2020}
\usepackage{bm}
\usepackage{times}
\usepackage{epsfig}
\usepackage{graphicx}
\usepackage{amsmath}
\usepackage{amssymb}
\usepackage{multirow}

\usepackage[pagebackref=true,breaklinks=true,letterpaper=true,colorlinks,bookmarks=false]{hyperref}

\usepackage[pagebackref=true,breaklinks=true,letterpaper=true,colorlinks,bookmarks=false]{hyperref}

\FGfinalcopy 

\IEEEoverridecommandlockouts                              
\def\FGPaperID{99} 

\title{\LARGE \bf
Taking Control of Intra-class Variation in Conditional GANs\\
Under Weak Supervision
}

\author{\parbox{16cm}{\centering
    {\large Richard Marriott$^{1,2}$, Sami Romdhani$^2$ and Liming Chen$^1$}\\
    {\normalsize
    $^1$ Ecole Centrale de Lyon, Ecully, France\\
    $^2$ IDEMIA, Courbevoie, France}}
}

\begin{document}

\ifFGfinal
\thispagestyle{empty}
\pagestyle{empty}
\else
\author{Anonymous FG2020 submission\\ Paper ID \FGPaperID \\}
\pagestyle{plain}
\fi
\maketitle

\begin{abstract}

Generative Adversarial Networks (GANs) are able to learn mappings between simple, relatively low-dimensional, random distributions and points on the manifold of realistic images in image-space. The semantics of this mapping, however, are typically entangled such that meaningful image properties cannot be controlled independently of one another. Conditional GANs (cGANs) provide a potential solution to this problem, allowing specific semantics to be enforced during training. This solution, however, depends on the availability of precise labels, which are sometimes difficult or near impossible to obtain, e.g. labels representing lighting conditions or describing the background. In this paper we introduce a new formulation of the cGAN that is able to learn disentangled, multivariate models of semantically meaningful variation and which has the advantage of requiring only the weak supervision of binary attribute labels. For example, given only labels of ambient / non-ambient lighting, our method is able to learn multivariate lighting models disentangled from other factors such as the identity and pose. We coin the method intra-class variation isolation (IVI) and the resulting network the IVI-GAN. We evaluate IVI-GAN on the CelebA dataset and on synthetic 3D morphable model data, learning to disentangle attributes such as lighting, pose, expression, and even the background.

\end{abstract}

\section{Introduction}\label{Introduction}

Generative Adversarial Networks (GANs) \cite{NIPS2014_5423} are capable of generating highly realistic synthetic data-samples. This capability is a direct result of the insightful training objective of the generative network: to produce synthetic data that cannot be distinguished from the real thing by a second adversarial network aiming to classify samples as either real or fake. Recent advances such as use of the Wasserstein loss \cite{arjovsky17a} and various forms of regularisation \cite{NIPS2017_7159, miyato2018spectral, brock2017neural} and training techniques \cite{karras2018progressive} mean that training of GANs is now stable enough to generate data-samples with thousands of dimensions. For example, deep convolutional GANs (DC-GANs) can now produce convincing images at mega-pixel resolution or higher \cite{karras2018progressive, Brock2018LargeSG}.

While classical DC-GANs can generate realistic images, the precise form of these images cannot be easily controlled. The subject of the images is dependent in some way on the values of the random input vector, but \textit{a priori} we do not know in what way. An obvious solution to this problem is the Conditional GAN (cGAN), in which the GAN's discriminator is trained to distinguish \textit{real image+label} pairs from \textit{fake image+label} pairs, thereby encouraging the generator to produce images that correctly correspond to the label upon which it is conditioned. In their typical form, cGANs are trained under the strong supervision of these labels: given binary category labels, images belonging to certain categories can be generated (but whose specific form is still governed by the entangled, real-valued random vector); given continuous, real-valued labels, cGANs can be forced to obey more specific semantics. Previous works, for example, have conditioned on binary hair-colour labels \cite{Choi_2018_CVPR_StarGAN}, and on continuous expression action unit magnitudes \cite{Pumarola2018GANimationAF}. However, the annotation of training images with descriptions of multi-dimensional phenomena such as expression and lighting, is notoriously difficult.

\begin{figure}[t]
\begin{center}
\includegraphics[width=0.95\linewidth]{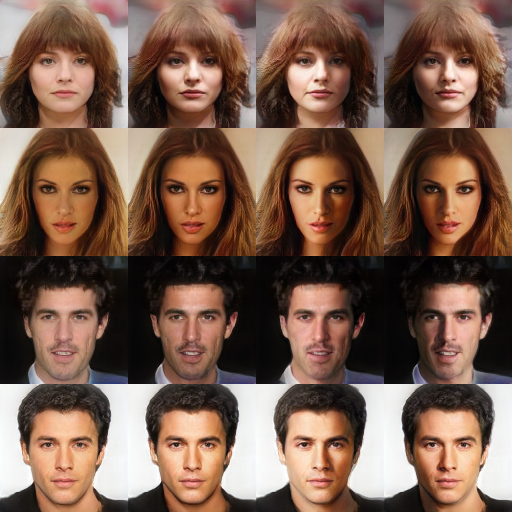}
\end{center}
\caption{Lighting conditions manipulated by IVI-GAN for four synthetic identities. Left-hand column: $\bm{\rho}_{lighting} = \mathbf{0}^4$ (ambient); the other columns show the effect of assigning a value of $3.0$ or $-3.0$ to individual elements of the lighting vector while keeping other parameters constant.}
\label{fig:IVI_lighting}
\end{figure}

In this paper we propose a new formulation of cGAN which is able to learn disentangled, multivariate representations of variation in images using only the weak supervision of simple, binary labels indicating the presence or absence of that form of variation, e.g. ``ambient lighting / non-ambient lighting''. Training data need not be labelled with precise parameter-estimations of prior models of the image property. We coin the method \textit{intra-class variation isolation} (IVI) and the resulting network the \textit{IVI-GAN}. A video demonstrating the continuous variation of various isolated parameter sets can be viewed at \href{https://youtu.be/hoWOFeADwdY}{https://youtu.be/hoWOFeADwdY}.

Our contributions are
\begin{itemize}
  \item An investigation of weakly supervised learning in conditional GANs and their ability to learn disentangled semantics given only binary labels;
  \item Proposition of a novel, physically motivated method of employing the GAN's generator which acts to enforce the separation of lighting from other forms of variation;
  \item A novel conditional GAN formulation that allows disentangled, multivariate models of variation to be learned with only the weak supervision of binary labels whilst preserving the identity: the IVI-GAN.
\end{itemize}

The rest of the paper is organised as follows: In Section \ref{Related_work} we discuss related work; in Section \ref{IVI-Learning} we formalise our novel conditional GAN formulation; in Section \ref{Implementation} we give specific details of our implementation; and in Section \ref{Results} we present an analysis of results produced using the IVI-GAN.


\section{Related Work}\label{Related_work}


Most recent works aiming to manipulate the semantics of synthesised images take the form of image-translation networks or auto-encoders with adversarial losses added to help ensure that images look realistic. In \cite{Emotion_preserving_face_frontalisation} face images are translated from being ``at pose'' to frontal while a discriminator ensures realism and that expression remains constant. The identity and other image properties, however, are only preserved via a pixel-wise comparison of the generated image with a ground-truth frontal image. Since such corresponding pairs of images are rarely available, other researchers have instead proposed auto-encoding methods such as CycleGAN \cite{CycleGAN}, whose image-shaped latent space is trained to resemble the distribution of a different class of images, not necessarily paired with the input image. For example, in both \cite{Choi_2018_CVPR_StarGAN} and \cite{Face_Synthesis_and_Domain_Adaptation}, a discriminator is used to encourage generated images to belong to different expression categories, while the cycle-consistency loss ensures that the general structure of the image remains unaffected, thus implicitly preserving the identity. In \cite{Pumarola2018GANimationAF}, continuous expression action unit labels are used to provide continuous control over the transformed images, rather than just control over the expression category. The downside, however, is that precise, real-valued labels are difficult to obtain. In addition, methods relying on cycle-consistency losses cannot be easily used to manipulate pose since this involves more significant alterations to image structure, thereby destroying the implicit constraint on identity.

Other works consider identity preservation explicitly by adding biometric losses. In \cite{Continuous_Emotion_Labels} a pre-trained biometric network is added as a constraint during manipulation of expression. In \cite{Bao_2018_CVPR} a biometric network is trained in parallel with the generative network whereas in \cite{Tran_2017_CVPR_DRGAN} the GAN's discriminator itself is trained to classify the identity as a secondary task. Both \cite{Bao_2018_CVPR} and \cite{Tran_2017_CVPR_DRGAN} are able to convincingly modify the pose of input images. All of these methods, however, require strong supervision to control image properties: in \cite{Tran_2017_CVPR_DRGAN} fine-grained pose-category labels are needed and in \cite{Continuous_Emotion_Labels}, similar to \cite{Pumarola2018GANimationAF}, continuous expression labels are needed.

InfoGAN \cite{InfoGAN} is the only work of which we are aware that attempts to disentangle control of image properties in an entirely unsupervised manner. By maximising the mutual information between a small subset of input parameters and the generated images, those parameters are attributed control over the most significant forms of variation in the image dataset. In practice these forms of variation tend to resemble semantics such as pose and lighting. However, there is no guarantee of which semantics will be learned, nor that identity will be preserved. Rather than an entirely unsupervised approach, in this work we propose a weakly supervised alternative, compromising between the unpredictable semantics of InfoGAN and the strong requirement for precise labels of typical conditional generation methods. Our weakly supervised technique allows the learning of specific forms of variation but does not require a prior model of that variation.

Concurrent with our own work, \cite{Interpreting_the_latent_space} demonstrates an alternative method for achieving continuous control over image properties using only binary labels. Rather than training a conditional GAN on labelled training images, labels are used retrospectively to identify the principal directions of variation of various attributes in the latent space of a pre-trained GAN. In their work, no special attention is paid to preserving identity. Also, in the absence of multiple labels, their method is not capable of identifying multi-directional forms of variation such as illumination or the configuration of the background. In Section \ref{Results} we add a comparison with \cite{Interpreting_the_latent_space} since, despite these differences, it is the work most closely related to our own.


\section{Intra-class variation isolation}\label{IVI-Learning}

Before describing intra-class variation isolation, we introduce the conditional GAN. Our best results are achieved using Wasserstein GANs and so we present this version.

\subsection{Conditional GANs}\label{cGAN}

A conditional GAN (cGAN) \cite{DBLP:journals/corr/MirzaO14} consists of two networks that are trained alternately: the discriminator, $D$, is trained to discriminate fake images from real ones; and the generator, $G$, is trained to generate images that will better fool the discriminator. The function that is minimised in order to train the discriminator is
\begin{multline}\label{eq: D_loss}
    \mathcal{L}_{\theta_D} = \mathbb{E}_{(\mathbf{x}, \mathbf{y})\sim p_{data}}[D(\mathbf{x}, \mathbf{y}; \theta_D)]\\
    - \mathbb{E}_{\mathbf{z}\sim\mathcal{N}, \bm{\rho}\sim p_{data}}[D(G(\mathbf{z}, \bm{\rho}; \theta_G), \bm{\rho}; \theta_D)]
\end{multline}
where $\mathbf{x}$ is an image and $\mathbf{y}\in\mathbb{R}^n$ the associated vector of labels drawn from the distribution of real data; $\mathbf{z}\in\mathcal{N}^m$ is a random Gaussian vector but could alternatively be drawn from any other real-valued distribution; $\bm{\rho}\in\mathbb{R}^n$ is a vector of labels used to condition the generator, here selected at random from the training dataset; and  $\theta_D$ and $\theta_G$ are the parameters of the discriminator and generator networks respectively. In the Wasserstein GAN, the output of $D$ is a single, unbounded scalar value; i.e. no \textit{squashing-function} is applied. Provided that $D$ remains $k$-Lipschitz continuous, minimising equation (\ref{eq: D_loss}) with respect to $\theta_D$ yields an estimator of the Wasserstein distance (to a scalar, $k$) between the distributions of real, labelled data and $\bm{\rho}$-parameterised, generated data \cite{arjovsky17a}. Lipschitz continuity is usually achieved by adding an additional \textit{gradient-penalty} term \cite{NIPS2017_7159} to the discriminator loss in equation (\ref{eq: D_loss}).

The trained discriminator is then used as the generator loss and equation (\ref{eq: G_loss}) is minimised with respect to $\theta_G$.
\begin{equation}\label{eq: G_loss}
    \mathcal{L}_{\theta_G} = \mathbb{E}_{\mathbf{z}\sim\mathcal{N}, \bm{\rho}\sim p_{data}}[D(G(\mathbf{z}, \bm{\rho}; \theta_G), \bm{\rho}; \theta_D)]
\end{equation}
Since generated images are never compared directly with the training images, the generator learns a \textit{smooth} distribution of realistic images from which new ones can be sampled.

\subsection{Intra-class variation isolation}\label{IVI-GAN}

In order to control some attribute of a generated image in a continuous fashion, for example the pose seen in a face-image, a typical cGAN requires each training image to be labelled with a precise and accurate pose-estimation. Our method, intra-class variation isolation (IVI), requires only the weak supervision of binary category labels indicating whether or not a particular form of variation is present. The IVI-GAN is then allowed to learn its own multivariate model of that form of variation.

Technically, the changes to the standard cGAN are very simple and involve modifying only the form of the labels provided to the cGAN. The function to be minimised to train the IVI-GAN's discriminator is
\begin{multline}\label{eq: IVI_D_loss}
    \mathcal{L}_{\theta_D} = \mathbb{E}_{(\mathbf{x}, \mathbf{y})\sim p_{data}}[D(\mathbf{x}, \mathbf{y}; \theta_D)] \\
    - \mathbb{E}_{\mathbf{z}\sim\mathcal{N}, \bm{\beta}\sim p_{data}}[D(G(\mathbf{z}, \bm{\rho}; \theta_G), \bm{\beta}; \theta_D)]
\end{multline}
where $\mathbf{y}\in\{0,1\}^n$ are now binary labels for $n$ categories, and $\bm{\beta}\in\{0,1\}^n$ are binary labels sampled from the same distribution as $\mathbf{y}$ (but, as before, do not necessarily have to be the same). The novel aspect of the loss is the way in which the random parameters, $\bm{\rho}$, are chosen.
\begin{equation}\label{eq: Parameter_function}
    \bm{\rho} = \begin{bmatrix}
    \mathbf{p}_1 \\
    \mathbf{p}_2 \\
    \vdots \\
    \mathbf{p}_n
    \end{bmatrix} \text{where } \mathbf{p_i} \in \begin{cases}
    \mathcal{N}^{q_i},& \text{if } \beta_i = 1\\
    \mathbf{0}^{q_i},& \text{if } \beta_i = 0\\
    \end{cases}
\end{equation}
Here, $\mathcal{N}^{q_i}$ is a random Gaussian vector of length $q_i$, and $\mathbf{0}^{q_i}$ is a vector of zeros. Analogous to the cGAN, the function then minimised to train the generator is
\begin{equation}\label{eq: IVI_G_loss}
    \mathcal{L}_{\theta_G} = \mathbb{E}_{\mathbf{z}\sim\mathcal{N}, \bm{\beta}\sim p_{data}}[D(G(\mathbf{z}, \bm{\rho}; \theta_G), \bm{\beta}; \theta_D)]
\end{equation}

\begin{figure*}[t]
\begin{center}
\includegraphics[width=0.6\linewidth]{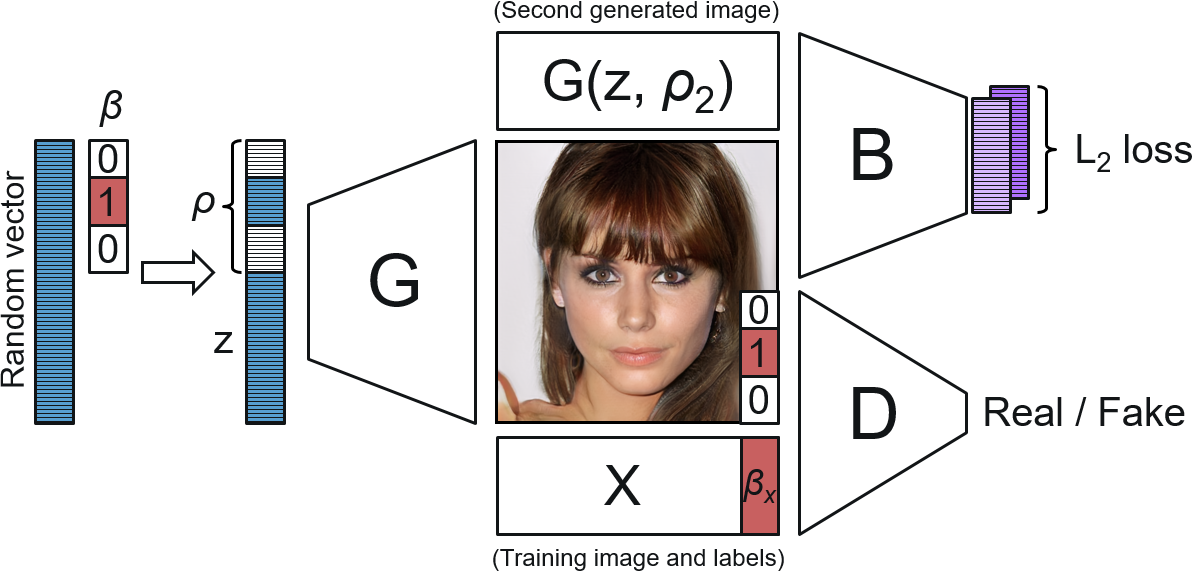}
\end{center}
\caption{An illustration of IVI-GAN. The real-valued parameter-vector, $\rho$, is formed by masking sections of an extended, random vector using the randomly selected, binary label-vector, $\beta$. It is $\beta$ (\textit{not} $\rho$) that is then fed to the discriminator with the generated image.}
\label{fig:IVI-GAN_diag}
\end{figure*}

N.B. The mechanism used to form $\bm{\rho}$ in (\ref{eq: Parameter_function}) could be interpreted as a function $f(\mathbf{z}', \bm{\beta}) = (\mathbf{z}, \bm{\rho})$, where $\mathbf{z}'$ is an extended random vector providing the additional random values used to form $\bm{\rho}$. Since this function could be implemented as part of a modified generator, $G'$ (which is permitted to have an arbitrary architecture), we have $G(\mathbf{z}, \bm{\rho}) = G(f(\mathbf{z}', \bm{\beta})) = G'(\mathbf{z}', \bm{\beta})$; i.e. equations (\ref{eq: IVI_D_loss}) and (\ref{eq: IVI_G_loss}) are mathematically equivalent to (\ref{eq: D_loss}) and (\ref{eq: G_loss}) but for binary labels. Figure \ref{fig:IVI-GAN_diag} depicts this intra-class variation isolation mechanism as part of our IVI-GAN.

All variation in images generated by the IVI-GAN must be derived from the combination of $\mathbf{z}$ and $\bm{\rho}$. The values of $\mathbf{z}$ are independent of $\bm{\beta}$ and so can always be used freely by the generator, irrespective of the labels. The parameter sets forming $\bm{\rho}$, however, are only available when certain forms of variation are labelled as being present. The idea is that the generator will then only use those parameters to describe that form of variation since the presence of non-zero parameters cannot be relied upon to describe anything else.

When labelling training images, it is best that $\beta_i = 0$ be used to describe a unique image property. For example, in the case of expression, $\beta_{exp} = 0$ should correspond to a neutral expression and $\beta_{exp} = 1$ to all non-neutral expressions. If $\beta_{exp} = 0$ were chosen to represent a non-unique property, such as ``not smiling'', then all of the different ways of not smiling would necessarily end up being encoded by $\mathbf{z}$. In the case of lighting, we chose $\beta_{lighting} = 0$ to represent ``ambient lighting''. This means that the colour and intensity of ambient light in our images is encoded by $\mathbf{z}$ but that all non-ambient lighting phenomena are encoded by $\bm{\rho}_{lighting}$.

We find that the IVI mechanism does indeed encourage disentanglement of labelled variation. This is aided by the natural parsimony of the generator which must find efficient ways of representing the training image distribution despite having only limited capacity. To do this, common features are, of course, reused by the generator to form different images. When making only subtle modification to images, e.g. adding lighting effects, the generator tends to leave the general structure of images unchanged, thus implicitly preserving the identity along with other image attributes. However, when making more significant changes to images, such as modifying the pose, this implicit identity preservation cannot be relied upon. We therefore introduce an explicit constraint on the identity which is described in the following section.

\subsection{The biometric identity-constraint}\label{Sec:Biometric_constraint}

To ensure that identity remains constant upon adjusting other image-properties, we have added a term to the generator loss of IVI-GAN involving a pre-trained biometric network \cite{hasnat2017deepvisage} that accepts a facial image as input and produces a 128-dimensional encoding of the identity.
\begin{equation}
    \mathcal{L}_{ID} = \mathbb{E}_{\mathbf{z}\sim\mathcal{N}}\left[ \left\lVert B(G(\mathbf{z}, \bm{\rho})) - B(G(\mathbf{z}, \bm{\rho}_2)) \right\rVert^2 \right]
\end{equation}\label{eq:L_ID}
where $B$ is the biometric network and $\bm{\rho}_2$ is a second set of random label-parameters. (N.B. we have dropped the $\theta$ for convenience of notation.) By running the generator twice for the same $\mathbf{z}$ but different $\bm{\rho}$, and constraining the identity encodings to remain close, we ensure that the identity is encoded as part of $\mathbf{z}$ and not affected by changes to $\bm{\rho}$. The full generator loss is then
\begin{equation}\label{eq: Full_G_Loss_ID}
    \mathcal{L}_{\theta_G} = \mathbb{E}_{\mathbf{z}\sim\mathcal{N}, \bm{\beta}\sim p_{data}}[D(G(\mathbf{z}, \bm{\rho}), \bm{\beta})] + \lambda_{ID} \mathcal{L}_{ID}
\end{equation}
where $\lambda_{ID}$ is a hyper-parameter to be tuned.

The biometric identity constraint is depicted on the right-hand side Figure \ref{fig:IVI-GAN_diag}.

\subsection{An additional, structural constraint for lighting}\label{Ave_lighting}

Despite the natural tendency of the generator to leave the structure of images unaffected when modifying properties such as lighting, subtle unwanted changes can still be seen, e.g. small changes to pose. To avoid this, we introduce a constraint on the image structure to be used when modifying lighting conditions.

Lighting is an additive phenomenon. For example, an image of a scene with two light sources is equivalent to the sum of two images of the same scene with the two light sources acting on it separately. One way to generate an image of a face under a particular lighting condition, therefore, is to add together two constituent images of the same subject. By re-formulating our generator in this way, the constraint that the composite image must appear realistic to the discriminator ensures that features in the two constituent images are of the same general structure. If not, the composite image would appear blurry and unrealistic due to misalignment and other inconsistencies. We propose that the generator be replaced with the following:
\begin{equation}\label{eq: G_ave}
    G(\mathbf{z}, \bm{\rho})_{comp} = G(\mathbf{z}, \mathbf{0}) + G(\mathbf{z}, \bm{\rho})
\end{equation}
where $\bm{\rho}$ represents lighting parameters only and $\mathbf{0}$ is a vector of zeros the same length as $\bm{\rho}$ indicating that ambient lighting should be generated. In our IVI-GAN, $G(\mathbf{z}, \bm{\rho})_{comp}$ replaces $G$ in both equation (\ref{eq: IVI_D_loss}) and (\ref{eq: IVI_G_loss}). By choosing $G(\mathbf{z}, \mathbf{0})$ as the second constituent image, we straightforwardly ensure that $G(\mathbf{z}, \mathbf{0})_{comp}$ generates ambiently lit images. We also find that this formulation can be used to help constrain facial structure when modifying the appearance of the background.


\section{Implementation}\label{Implementation}

Our implementation is built upon the stable and efficient progressive GAN of \cite{karras2018progressive}, a Tensorflow implementation of which was made publicly available by Nvidia. The progressive GAN begins by generating images of $4\times 4$ resolution and then progressively fades in new convolutional upscaling layers until the desired resolution is reached. There has been much recent work on improving the quality of GAN-generated images published in the literature. We tested a selection of these enhancements and found that our best results were produced by a progressive Wasserstein GAN with the standard gradient penalty (GP) term of \cite{NIPS2017_7159} where the weight of the GP term was allowed to evolve throughout training based on an \textit{adaptive lambda} scheme similar to that in \cite{Chen_2018_CVPR}. In the generator we use orthogonal initialisation of weights and replace the pixel-wise feature normalisation used in \cite{karras2018progressive} with the orthogonal regularisation of \cite{brock2017neural} using the suggested weight of 0.0001.

\subsection{Conditioning the GAN}\label{Conditioning}

The way in which labels and label-parameters are used to condition GANs is an open area of research. For example, \cite{miyato2018cgans_projectionD} finds that, given certain assumptions about the form of the distribution of data, the optimal method of conditioning the discriminator should be to learn some inner-product of the label-vector with the channels at each pixel; in \cite{Conditional_instance_norm} the generator network is conditioned via instance-normalisation parameters. We have used the more straight-forward method of concatenating label-vectors with inputs but expect that these more sophisticated methods of conditioning may be used to improve results. We concatenate our IVI parameter vector, $\bm{\rho}$, with the random vector on input to the generator, and on input to the discriminator we concatenate the binary labels, $\mathbf{y}$ and $\bm{\beta}$, as additional channels repeated at each pixel of the real and generated images respectively.

An analysis of where best to concatenate conditioning vectors was performed in \cite{DBLP:journals/corr/PerarnauWRA16}. They predicted that earlier in the network should be better since the discriminator would be afforded more learning interactions with the information. In fact they conclude that it is best to concatenate the conditioning with the first hidden layer. However, in a progressive architecture, the first convolutional layer of the discriminator is not faded in until the final stages of training. In our case, it therefore makes more sense to concatenate the conditioning with the image where it is then scaled down and fed to each layer of the progressive network as they are being faded in. Ultimately, once all layers of the progressive discriminator are active, the conditioning information and image are only fed to the first layer of the discriminator.

\begin{figure}[t]
\begin{center}
\includegraphics[width=0.8\linewidth]{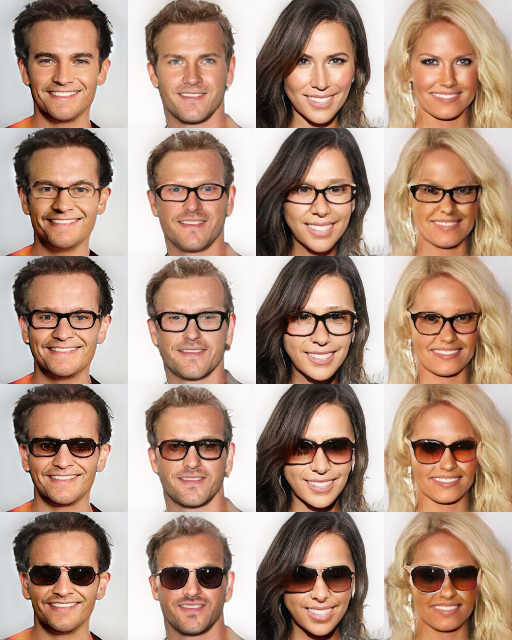}
\end{center}
\caption{A continuous eye-wear model learned by IVI-GAN. The style of eye-wear is controlled by the direction of a two-dimensional unit vector. Setting the length of the vector to zero removes the eye-wear (top row).}
\label{fig:Glasses_model}
\end{figure}

\begin{figure}[t]
\begin{center}
\includegraphics[width=0.8\linewidth]{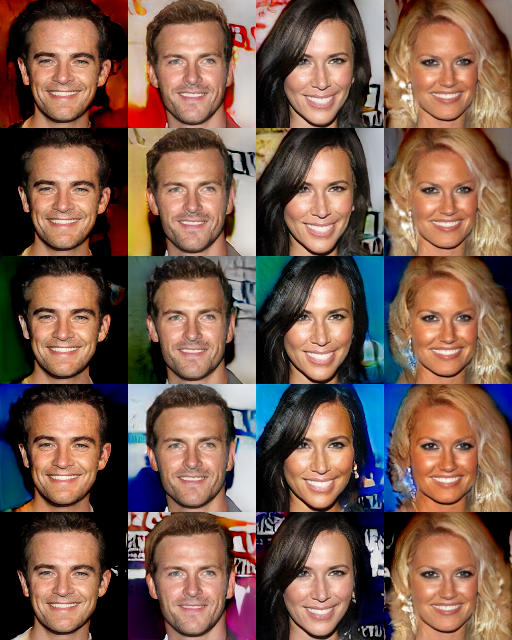}
\end{center}
\caption{A colourful selection of images demonstrating the capability of IVI-GAN to generate a range of different backgrounds while preserving identity and other image attributes.}
\label{fig:Background_model}
\end{figure}

\begin{figure}[t]
\begin{center}
\includegraphics[width=0.6\linewidth]{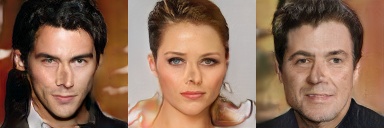}
\end{center}
\caption{Examples of broken nose features generated during tests of an auxiliary classifier (not used in IVI-GAN).}
\label{fig:Broken_noses}
\end{figure}

Many applications make use of auxiliary classifiers (ACs) \cite{pmlr-v70-odena17a} as a way of ensuring that conditional parameters are not ignored during generation of images. We tested this method in conjunction with IVI-GAN using the auxiliary classifier already implemented in Nvidia's progressive GAN code. However, we found results to be unsatisfactory. As noted in \cite{miyato2018spectral}, auxiliary classifiers encourage the generator to produce images that are easy to classify; a goal which is not in alignment with the principal training objective of the GAN. We found that, given a large weight in the discriminator loss, the AC-term caused mode-collapse, squeezing variation into narrow, well-separated categories. For example, upon varying continuous, conditional pose parameters we observed a discrete jump in the generated pose between frontal and large poses. Giving less weight to the AC-term ameliorated the discrete jumps in pose. However, more subtle artefacts remained, such as broken noses pointing in one direction or the other; a feature obviously used by the discriminator to help classify slightly non-frontal poses. See Figure \ref{fig:Broken_noses} for examples of this behaviour.

\section{Results}\label{Results}

We have evaluated IVI-GAN on the CelebA dataset \cite{CelebA}, and on a dataset of synthetic face images generated using the Basel 3D morphable model (3DMM) \cite{blanz1999morphable}. In Section \ref{Sec. CelebA} we present the results for CelebA, including a qualitative comparison with similar results taken from \cite{Interpreting_the_latent_space}. In Section \ref{Sec. Weak_learning} we investigate pose changes in images generated from CelebA to give an idea of the form and consistency of multivariate models learnt via weak supervision. We then show results demonstrating the efficacy of our biometric identity constraint in Section \ref{Sec. Biometric_efficacy} and finally, in Section \ref{Sec. 3DMM_results}, we show additional results for a balanced dataset of synthetic 3DMM images.

\begin{figure}[t]
\begin{center}
\includegraphics[width=1.0\linewidth]{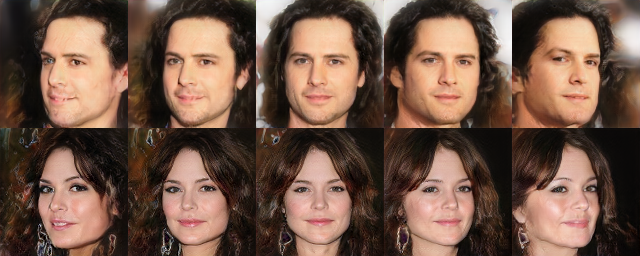}
\end{center}
\caption{Images demonstrating the effect of varying one of the pose parameters, used by IVI-GAN to represent yaw-like variation. The parameter is varied between $-3.0$ and $3.0$. In the middle column $\bm{\rho}_{pose} = (0, 0)$.}
\label{fig:pose_yaw}
\end{figure}

\begin{figure}[t]
\begin{center}
\includegraphics[width=1.0\linewidth]{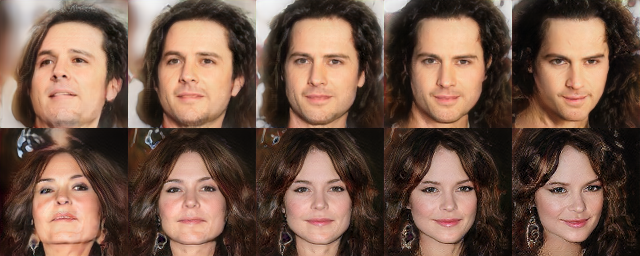}
\end{center}
\caption{Images demonstrating the effect of varying the other pose parameter, used by IVI-GAN to represent pitch-like variation. The parameter is varied between $-3.0$ and $3.0$. In the middle column $\bm{\rho}_{pose} = (0, 0)$.}
\label{fig:pose_pitch}
\end{figure}

\subsection{Taking control of variation in CelebA}\label{Sec. CelebA}

We trained IVI-GAN on a 100k image subset of the CelebA dataset. Images were prepared in a similar way to the CelebA-HQ dataset of \cite{karras2018progressive} but super-resolution was not used. Our network was trained progressively up to a resolution of $128 \times 128$ and was conditioned on a selection of the attribute labels available with CelebA. These attributes were complemented with binary labels for lighting, the background, and for pose. Lighting and background labels were found by hand labelling a set of 10k images as containing either ambient or non-ambient lighting, and having either plain or busy/coloured backgrounds. Two simple classifiers were then trained to label the remaining images. Pose labels were found by applying an off-the-shelf pose detector and categorising all images with yaw and pitch greater than three degrees as being non-frontal.

Figures \ref{fig:IVI_lighting}, \ref{fig:Glasses_model}, \ref{fig:Background_model}, \ref{fig:pose_yaw} and \ref{fig:pose_pitch} show the effect of varying the multi-dimensional parameter vectors learned for lighting (4 parameters), eye-wear (2 parameters normalised to unit length), background (10 parameters) and pose (2 parameters). Each row of images in Figure \ref{fig:Glasses_model} corresponds to a particular configuration of $\bm{\rho}_{glasses}$. We use a vector of two parameters normalised to unit length. Only four instances of variation are shown but the style of glasses can be varied continuously by rotating the unit vector, with each style morphing smoothly into the next. Glasses can be removed completely by setting $\bm{\rho}_{glasses} = (0, 0)$ (top row). We see that modifications to the style of glasses are well disentangled from the other image parameters and from the identity.

In Figure \ref{fig:Background_model}, each row corresponds to a different random instantiation of $\bm{\rho}_{background}$. (Setting $\bm{\rho_{background} = \mathbf{0}^{10}}$ results in the same set of images as shown in the first row of Figure \ref{fig:Glasses_model}.) Again, we see that modifications to the background leave other image properties and the identity largely unaffected. However, we notice that certain features of the background have a tendency to be present in most images of certain identities. We believe this effect is due to unwanted, spurious correlations in the training dataset.

\begin{figure}[t]
\begin{center}
\includegraphics[width=1.0\linewidth]{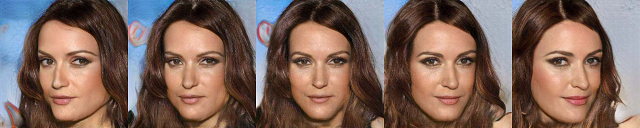}
\includegraphics[width=0.8\linewidth]{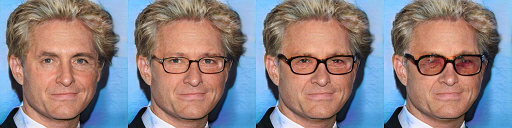}
\end{center}
\caption{Images taken from Shen et. al. (2019) \cite{Interpreting_the_latent_space}.}
\label{fig:interpreting_latent_space_pose}
\end{figure}

Although unguided by precise labels, Figures \ref{fig:pose_yaw} and \ref{fig:pose_pitch} show that IVI-GAN has learned to use one pose parameter to represent yaw-like variation and the second to represent pitch-like variation. Other image properties such as lighting, the background and the identity remain consistent. These results (and also those of Figure \ref{fig:Glasses_model}) can be compared with those in Figure \ref{fig:interpreting_latent_space_pose} which have been taken from Shen et. al. (2019) \cite{Interpreting_the_latent_space}. Note that in \cite{Interpreting_the_latent_space}, images were generated at higher resolution. Here, we have down-sampled them to $128 \times 128$ resolution for closer comparison with our own. The identities in Figure \ref{fig:interpreting_latent_space_pose} seem to be reasonably well preserved despite \cite{Interpreting_the_latent_space} having taken no explicit steps to achieve this. We suspect that this would not be the case, however, if pitch were to be varied using the same method. With fewer images in the training set exhibiting pitch variations, correlation of large pitches with certain identities can lead to shifts towards those identities. In contrast to \cite{Interpreting_the_latent_space}, IVI-GAN simultaneously learns a multivariate representation of both yaw \textit{and} pitch. It also explicitly ensures that identity-drift is kept to a minimum via its biometric identity constraint.

\subsection{Weak learning of multivariate models}\label{Sec. Weak_learning}

\begin{figure}[t]
\begin{center}
\includegraphics[width=0.95\linewidth]{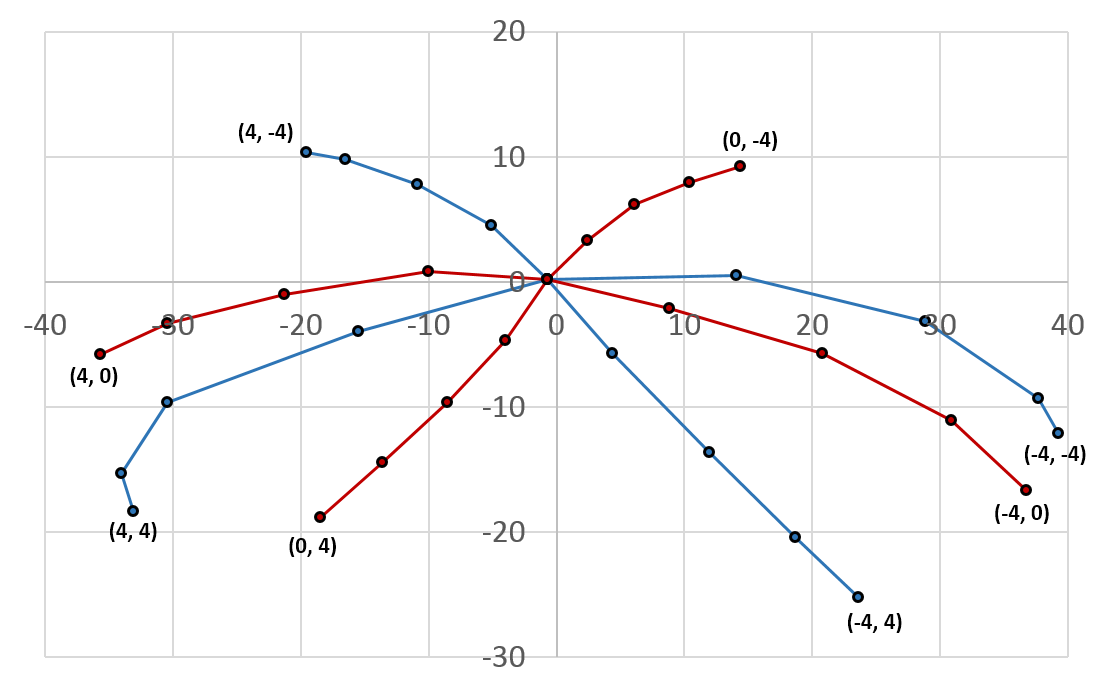}
\end{center}
\caption{Detected poses in images generated by an IVI-GAN. Horizontal and vertical axes indicate detected yaw and pitch for a selection of parameter values (indicated in the plot) averaged over 100 identities.}
\label{fig:Spider_main}
\end{figure}

\begin{figure}[t]
\begin{center}
\includegraphics[width=1.0\linewidth]{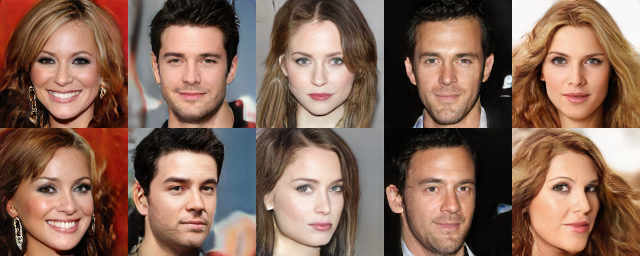}
\end{center}
\caption{Examples of images analysed in Table \ref{tab:Pose_consistency} generated by an IVI-GAN. Five random identities are shown in frontal poses (top row) and with pose parameters prescribed as $\rho_{pose} = (0.0, 2.0)$ (bottom row).}
\label{fig:IVI_GAN_pose_consistency}
\end{figure}

\begin{figure}[t]
\begin{center}
\includegraphics[width=1.0\linewidth]{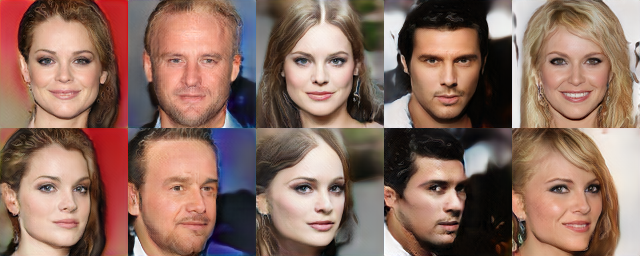}
\end{center}
\caption{Examples of images analysed in Table \ref{tab:Pose_consistency} generated by the cGAN. Five random identities are shown in frontal poses (top row) and with pose parameters prescribed as $\rho_{pose} = (23.8^{\circ}, -6.5^{\circ})$ (bottom row).}
\label{fig:CC_GAN_pose_consistency}
\end{figure}

IVI-GAN is able to learn its own models of variation given only binary labels that indicate the presence or absence of that form of variation. A desirable property of such a model is that the same parameter values result in the same semantic properties irrespective of the identity and other image properties. In Table \ref{tab:Pose_consistency} we compare the consistency of poses generated by IVI-GAN (without ID constraint) with those generated by a cGAN conditioned on precise, real-valued pose labels. The only difference between the cGAN and the IVI-GAN is that, for the cGAN, $\bm{\rho}_{pose}$ are pose labels instead of random parameters, and are fed to the discriminator in place of the binary labels, $\beta_{pose}$. We generated $100$ random identities for each of the pose parameter configurations given in Table \ref{tab:Pose_consistency} and then used a pose-detector to find the mean generated pose and the standard deviation of angles from that direction. Note that, for a more straightforward comparison, the pose parameter configurations that were fed to the cGAN are the mean poses detected in the images generated by IVI-GAN. We see that, despite the absence of strong supervision, IVI-GAN is able to generate poses with a consistency close to that of a standard cGAN. (Note that the consistency of the cGAN statistics may be artificially high since the same detector was used to label the training images.) The form of the pose model learned by IVI-GAN (with ID constraint) is depicted in Figure \ref{fig:Spider_main} and examples of the images analysed in Table \ref{tab:Pose_consistency} are given in Figures \ref{fig:IVI_GAN_pose_consistency} and \ref{fig:CC_GAN_pose_consistency}. It can be seen that the visual quality of images generated by IVI-GAN is similar to that of the cGAN.

\begin{table}[t]
\caption{Statistics of poses detected in images of 100 random identities for the given parameter-configurations.}
\label{tab:Pose_consistency}
\begin{center}
\begin{tabular}{|l|c|r|r|c|}
\hline
\multicolumn{2}{|c|}{\textbf{GAN configuration}} & \multicolumn{2}{|c|}{\textbf{Mean pose}} & \multicolumn{1}{|c|}{\textbf{StdDev}} \\ \cline{1-4}
\textbf{Type} & $\bm{\rho}$ & \multicolumn{1}{|c|}{\textbf{yaw}} & \multicolumn{1}{|c|}{\textbf{pitch}} & \multicolumn{1}{|c|}{\textbf{pose}} \\ \hline
\hline
\multirow{3}{*}{IVI-GAN} & $(0.0, 1.0)$ & $10.2^{\circ}$ & $-1.3^{\circ}$ & $7.5^{\circ}$ \\ \cline{2-5}
 & $(0.0, 2.0)$ & $23.8^{\circ}$ & $-6.5^{\circ}$ & $8.8^{\circ}$ \\ \cline{2-5}
 & $(0.0, 3.0)$ & $33.7^{\circ}$ & $-12.6^{\circ}$ & $8.8^{\circ}$ \\
 \hline
\multirow{3}{*}{cGAN} & $(10.2^{\circ}, -1.3^{\circ})$ & $9.3^{\circ}$ & $0.8^{\circ}$ & $4.8^{\circ}$ \\ \cline{2-5}
 & $(23.8^{\circ}, -6.5^{\circ})$ & $23.5^{\circ}$ & $-5.2^{\circ}$ & $6.3^{\circ}$ \\ \cline{2-5}
 & $(33.7^{\circ}, -12.6^{\circ})$ & $32.5^{\circ}$ & $-11.4^{\circ}$ & $6.9^{\circ}$ \\ \hline
\end{tabular}
\end{center}
\end{table}

\subsection{Efficacy of the biometric constraint}\label{Sec. Biometric_efficacy}

\begin{figure}[t]
\begin{center}
\includegraphics[width=0.45\linewidth]{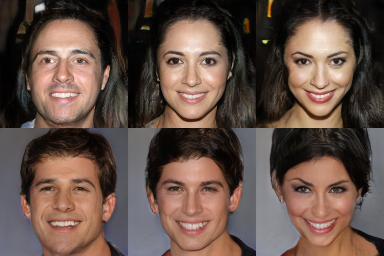}
\end{center}
\caption{Results demonstrating drift in identity when varying a pitch-like parameter in IVI-GAN \textit{without} biometric identity constraint. Each row of images was generated from the same $\mathbf{z}$ vector.}
\label{fig:ID_problems}
\end{figure}

As previously mentioned, modifications to images that significantly affect their general structure, such as changes to pose, can lead to identity shift and shifts in other image properties. Changes to the property of pitch seem to be particularly prone to this issue. (See Figure \ref{fig:ID_problems}.) To counter these problems, IVI-GAN incorporates the explicit, biometric identity constraint described in Section \ref{Sec:Biometric_constraint}. The biometric network \cite{hasnat2017deepvisage} was pre-trained on images of resolution $96 \times 96$ and so we only activate the additional ID-loss during the final stabilisation period of the training of the progressive GAN. We performed experiments with $\lambda_{ID} = [1.0, 0.1, 0.01, 0.001, 0.0001]$ and finally use $\lambda_{ID} = 0.0001$. Higher values were found to inhibit pose-variation too much.

Comparisons were made between biometric encodings of $100$ random identities generated at a frontal pose and the same identities (same $\mathbf{z}$ vectors) generated using the pose parameter configurations given in Table \ref{tab:ID_consistency}. The final three rows of Table \ref{tab:ID_consistency} show that $\mathcal{L}_{ID}$ has the desired effect of ensuring that images generated at larger poses match their frontal counterparts. With the ID-constraint enabled, the lowest mean detected cosine similarity is now $0.78$ at an average pose of $(-13.6^{\circ}, -14.4^{\circ})$. Note that the biometric network used to produce these matching scores was not the same as that used to constrain the IVI-GAN. Figures \ref{fig:pose_yaw} and \ref{fig:pose_pitch} show examples of the extent to which identity is constrained as pose parameters are varied.

\begin{table}[t]
\caption{Identity-matching statistics for images generated at pose vs. frontal ($\bm{\rho} = (0.0, 0.0)$). Matchings considered to be failures ($< 0.7$) are indicated in red.}
\label{tab:ID_consistency}
\begin{center}
\begin{tabular}{|l|c|c|c|}
\hline
\multicolumn{2}{|c|}{\textbf{GAN configuration}} & \multicolumn{1}{|c|}{\textbf{Cosine}} & \multicolumn{1}{|c|}{\textbf{StdDev}} \\ \cline{1-2}
\textbf{Type} & $\bm{\rho}$ & \multicolumn{1}{|c|}{\textbf{similarity}} & \multicolumn{1}{|c|}{\textbf{cosine sim.}} \\ \hline
\hline
\multirow{3}{*}{IVI} & $(0.0, 1.0)$ & $0.83$ & $0.19$ \\ \cline{2-4}
 & $(0.0, 2.0)$ & \textcolor{red}{$0.66$} & $0.17$ \\ \cline{2-4}
 & $(0.0, 3.0)$ & \textcolor{red}{$0.54$} & $0.11$ \\
 \hline
 \multirow{3}{0.1em}{IVI\\+$\mathcal{L}_{ID}$} & $(0.0, 1.0)$ & $0.94$ & $0.10$ \\ \cline{2-4}
 & $(0.0, 2.0)$ & $0.86$ & $0.16$ \\ \cline{2-4}
 & $(0.0, 3.0)$ & $0.78$ & $0.17$ \\ \hline
\end{tabular}
\end{center}
\end{table}

\subsection{Learning from a balanced, synthetic dataset}\label{Sec. 3DMM_results}

\begin{figure}[t]
\begin{center}
\includegraphics[width=0.75\linewidth]{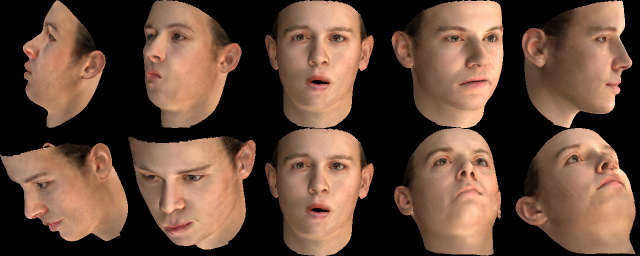}
\end{center}
\caption{Results demonstrating pose-variation in images generated by IVI-GAN trained on a synthetic dataset. The two rows show the effect of varying the two, uniform pose parameters between $-3.0$ and $3.0$. All other parameters were kept the same.}
\label{fig:3dmm_poses}
\end{figure}

\begin{figure}[t]
\begin{center}
\includegraphics[width=0.75\linewidth]{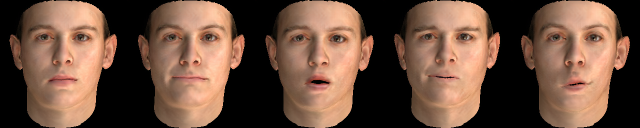}
\includegraphics[width=0.75\linewidth]{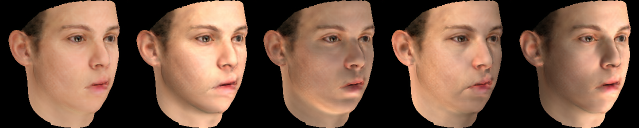}
\end{center}
\caption{Results demonstrating expression and lighting variation in images generated by IVI-GAN trained on a synthetic dataset. The left-hand images show neutral expression ($\bm{\rho}_{exp} = \mathbf{0}^8$, top) and ambient lighting ($\bm{\rho}_{lighting} = \mathbf{0}^9$, bottom). The other images show the effects of activating individual, expression and lighting parameters with values of $-3.0$ or $3.0$.}
\label{fig:3dmm_expression_and_lighting_modes}
\end{figure}

The quality of generated images at large poses and containing other extreme conditions is limited by the availability of such images in training datasets. As a cleaner test of IVI, we trained IVI-GAN on synthetic face images generated by a 3D morphable model (3DMM) \cite{blanz1999morphable} and lit using a spherical harmonic lighting model \cite{Ramamoorthi9SH}. Identities and expressions were sampled from random Gaussian distributions, and lighting and pose from uniform distributions. Figure \ref{fig:3dmm_poses} demonstrates that, given adequate data, IVI-GAN is able to generate high-quality results for the full range of poses, i.e. for yaws in the range $[-90^{\circ}, 90^{\circ}]$ (top) and pitches in the range $[-45^{\circ}, 45^{\circ}]$ (bottom). We note that, since there is no explicit constraint on expression, the expression we selected for the frontal image is lost during the disruptive pose changes. The desired expression can often be recovered, however, by readjusting expression parameters afterwards, as has been done in the bottom row of Figure \ref{fig:3dmm_expression_and_lighting_modes}. Since adjusting the expression is a more subtle image-modification, it does not affect the pose.

With full control over the synthetic dataset, we were able to easily generate labels of neutral/non-neutral expression. (Similar labels were not available during our tests on CelebA.) The top row of Figure \ref{fig:3dmm_expression_and_lighting_modes} shows the range of expressions learnt by IVI-GAN, whilst the bottom row shows some of the more distinctive lighting modes that were learned. Note that the lighting condition remains consistent as expression is varied and vice-versa.

\section{Conclusions}\label{Conclusion}

We have shown that it is possible to adapt a conditional GAN in order to gain continuous, disentangled control over image attributes without the need for extensive labelling. Only simple binary labels, indicating whether an attribute is present in one form or another, are required. The GAN is then allowed to discover its own, multivariate way of modelling the variation present within that attribute category in an unsupervised fashion. To the best of our knowledge, IVI-GAN is the first network to achieve this separation in a weakly supervised manner. In conjunction with an explicit identity constraint and a further, physically motivated image-structure constraint used when manipulating lighting and the background, IVI has been shown to be highly effective via evaluation on CelebA and on a synthetic dataset of 3DMM images. Potential applications of IVI-GAN include dataset-augmentation and image-manipulation.


\end{document}